\documentclass[conference]{IEEEtran}
\usepackage{amsmath,amssymb,amsthm}
\usepackage{cite}
\usepackage{algorithm}
\usepackage{algorithmic}
\usepackage{caption}
\usepackage{ltablex}
\usepackage{booktabs}
\usepackage{hyperref}
\usepackage{subcaption}
\usepackage{url}
\usepackage{color}
\usepackage{graphicx}
\usepackage{array}
\usepackage{multicol}
\usepackage{caption}
\usepackage{booktabs}
\usepackage{float}
\usepackage{etoolbox}

\usepackage{bm}

\usepackage[bottom]{footmisc}

\usepackage{array}
\newcolumntype{L}[1]{>{\raggedright\let\newline\\\arraybackslash\hspace{0pt}}m{#1}}
\newcolumntype{C}[1]{>{\centering\let\newline\\\arraybackslash\hspace{0pt}}m{#1}}
\newcolumntype{R}[1]{>{\raggedleft\let\newline\\\arraybackslash\hspace{0pt}}m{#1}}
\newcommand{\bx}{\ensuremath{\mathbf{x}}}

\newcommand{\bz}{\ensuremath{\mathbf{z}}}
\newcommand{\bw}{\ensuremath{\mathbf{w}}}
\newcommand{\ba}{\ensuremath{\mathbf{\alpha}}}
\newcommand{\bb}{\ensuremath{\pmb{\beta}}}
\DeclareMathOperator*{\argmin}{arg\,min}
\newtheorem{proposition}{Proposition}
\newtheorem{definition}{Definition}

\definecolor{dgreen}{RGB}{63, 175, 115}

\begin{document}
\sloppy





%

\title{A budget-constrained inverse classification framework for smooth classifiers}

\author{\IEEEauthorblockN{Michael T. Lash\IEEEauthorrefmark{1}, Qihang Lin\IEEEauthorrefmark{2}, W. Nick Street\IEEEauthorrefmark{2} and Jennifer G. Robinson\IEEEauthorrefmark{3}}
	\IEEEauthorblockA{\IEEEauthorrefmark{1}Department of Computer Science, \IEEEauthorrefmark{2}Department of Management Sciences, \IEEEauthorrefmark{3}Department of Epidemiology\\
		The University of Iowa\\
		Iowa City, Iowa 52242\\
		\{michael-lash, qihang-lin, nick-street, jennifer-g-robinson\}@uiowa.edu}
}

\maketitle

\begin{abstract}

Inverse classification is the process of manipulating an instance
such that it is more likely to conform to a specific class. Past
methods that address such a problem have shortcomings. Greedy methods
make changes that are overly radical, often relying on data that is
strictly discrete. Other methods rely on certain data points, the
presence of which cannot be guaranteed. In this paper we propose a
general framework  and method that overcomes these and other limitations. The formulation of our method can use any
differentiable classification function. We demonstrate the method by using logistic regression and Gaussian kernel SVMs. We constrain the inverse
classification to occur on features that can actually be changed,
each of which incurs an individual cost. We further subject such changes
to fall within a certain level of cumulative change (budget). Our framework can also accommodate the estimation of (indirectly changeable) features
whose values change as a consequence of actions taken. Furthermore,
we propose two methods for specifying feature-value ranges that result
in different algorithmic behavior. We apply our method, and a proposed sensitivity analysis-based benchmark method, to two freely available datasets: Student Performance from the UCI Machine Learning Repository and a real-world cardiovascular disease dataset. The results
obtained demonstrate the validity and benefits of
our framework and method.

\end{abstract}




\section{Introduction}

In many predictive modeling problems, we are concerned less with the
actual prediction, and more with how an individual prediction might be
changed. Classification problems such as loan screening and college
admission have one output class that is clearly ``desired'' by a
test case. A person turned down for a loan would naturally wonder why
the decision was made, and more importantly, what they could do to
change the outcome on the next attempt. We use the term {\it inverse
	classification} to refer to the process of finding an optimal
set of changes to a test point so as to maximize its predicted probability
of the desired class label.

Problems such as this are prevalent in personalized medicine settings.
Consider, for example, lifestyle choices that minimize Patient 15's
long-term risk of cardiovascular disease (CVD) -- a randomly selected patient from our experiments in Section IV. An initial risk prediction,
estimated to be 32\%, is obtained using a trained, nonlinear
classifier, based on Patient 15's EHR data.
 With Patient
15's initial risk now known, we wish to work ``backwards'' through the
classifier to obtain recommendations that minimize the probability of CVD. We approach the
recommendation step by defining an optimization problem: what is the
smallest (or easiest) set of feasible changes that this person can
make in order to minimize the predicted probability of developing CVD?


Our first contribution in this work is to define an inverse classification
framework that produces realistic recommendations. We do so by
first partitioning features into two categories: unchangeable and changeable.
It would be impossible for Patient 15 to reduce her age -- this is an unchangeable
feature.  Changeable features are further partitioned into directly and indirectly
changeable categories.  Directly changeable features are immediately actionable --
we can recommend that Patient 15 adjust her diet, for example.  Indirectly changeable
features change as a consequence of manipulations to the directly changeable features,
but are themselves not actionable.  Blood glucose changes as Patient 15's diet is
altered, but cannot be directly altered itself.

In our framework, directly changeable features incur individual, attribute-wise cost.
 Cumulative costs across such features are constrained to be within a budgetary level.
 These costs and budget can be specified by either a domain expert, the individual (e.g.,~Patient 15), or some
 combination of the two.

The second contribution of this work is a method that solves the inverse classification problem within the specified framework. Our method uses the gradient information of classifiers to provide recommendations that minimize the probability of an undesirable class. Using such
a method within the specified framework we are able to provide recommendations
that reduce Patient 15's probability of CVD from 32\% to 3\%.


The third contribution we identify is to specify two bound-setting methods, Elastic and Hard-line, that
operate within the outlined framework allowing inverse classification to occur more freely or more rigidly, depending upon the problem. Lastly, we incorporate an indirect feature estimator, that adjusts features that change as a consequence of the directly alterable set of features.

In the remainder of the paper we discuss past work (Section II), our proposed framework and new method of inverse classification (Section III), our 16 experiments, conducted on two freely available datasets using our method and a sensitivity analysis-based benchmark method (Section IV), and the conclusions we make following these experiments (Section V).





\section{Related Work}

Inverse classification can be seen as a form of sensitivity analysis,
the process of examining the input features'
effects on the target output. While there are many forms of sensitivity analysis \cite{isukapalli1999,Yao2003}, inverse classification is most similar to local sensitivity analysis and variable perturbation method. Later on (Section III), we propose a benchmark method that is based on these.

Past works on inverse classification can be looked at from three perspectives: the manner in which the algorithm operates, the type of data the algorithm operates on, and the framework that guides the process of obtaining recommendations. Algorithm operation, which represents the optimization method employed, can be broken down into two groups: \textbf{greedy} \cite{Aggarwal2010,Chi2012,Yang2012,Mannino2000} and \textbf{nongreedy} \cite{Barbella2009,Pendharkar2002}. Greedy methods tend to focus on extreme objectives, which may not be realistic in the real world, while nongreedy methods tend to focus on more moderate objectives. This work uses the latter.

Algorithmic data types, which refers to the type of data a particular optimization algorithm has the capability of operating on, also fall into two categories: \textbf{discrete} \cite{Aggarwal2010,Chi2012,Yang2012} and \textbf{continuous} \cite{Mannino2000,Barbella2009,Pendharkar2002}. Discrete data types lead to coarse-grained recommendations, while continuous data types provide those that are more fine-grained. In this work, we focus on the latter, as precision recommendations are the goal.

Framework refers to the constraints that govern recommendation feasibility. These are manifested in the literature as either \textbf{unconstrained} \cite{Aggarwal2010,Chi2012,Yang2012} or \textbf{constrained} \cite{Barbella2009,Pendharkar2002,Mannino2000}. Unconstrained problems lead to unrealistic recommendations that may also be very extreme (e.g.,~`reduce your age by 30 years'). Constrained frameworks lead to more moderate and realistic recommendations. However, while \cite{Barbella2009,Pendharkar2002} focus on moderate objectives, they do not consider (1) what can/cannot be changed, (2) how hard it might be to change and (3), cumulatively, how willing someone may be to make changes. In \cite{Mannino2000} the authors consider (2), but do not consider (1) and (3). Additionally, in \cite{Barbella2009}, the formulation of \textit{border classification} relies on data points which lie exactly on the separating hyperplane; there is not guarantee that such points exist in practice. In this work we propose a framework that considers (1), (2) and (3).

\textcolor{black}{Inverse classification is a utility-based data mining topic and is thereby related to the subtopics of strategic \cite{Boylu2010} and adversarial \cite{Lowd2005} learning. In these topics it is assumed that a strategic agent may attempt to \textit{game} a learned classifier in order to conform to a desired class. Classifiers are then constructed taking such behavior into account. Such considerations do not need to be made in an inverse classification setting, however, as the goal is to provide explicit instructions to an intelligent agent (e.g.,~person) on how they can conform to a desired class, thereby making such accounts both unnecessary and undesirable.}

\section{An Inverse Classification Framework and Method}


In this section we propose a new inverse classification framework, and a method that can be used within the framework to solve the problem. We begin by generally discussing the problem and introducing some notation.

Suppose $\{(\bx^i,y^i)\}_{i=1,2,\dots,n}$ is a dataset of $n$, assumed to have been drawn i.i.d.~from some population distribution $\mathcal{P}$, where $\bx^i\in\mathbb{R}^{p}$ is a column feature vector of length $p$ and $y^i \in \{-1,1\}$ is the binary label associated with $\bx^i$ for $i=1,2,\dots,n$. Let $X = [\bx^1,...,\bx^n]^T \in \mathbb{R}^{n\times p}$ denote the matrix of training instances with $(\bx^i)^T$'s being its rows. Any number of classification models can be trained with this dataset and used to predict the class of new instances. Unlike typical classification settings, however, given a new instance $\bx\in\mathbb{R}^{p}$, our goal is not only to classify it to the positive or the negative class but also to recommend an update on $\bx$ that minimizes the probability of $\bx$ being classified as positive. We assume one unit change in each feature of $\bx$ will incur a cost and that only a limited amount of budget $B$ is available. We propose a numerical framework and algorithm that recommends an optimal change on $\bx$ based on a classification model that incorporates this budgetary constraint.

\subsection{Framework}

Suppose we are allowed to change some of the features of instance $\bx$  to obtain a new version $\bx'$. Also suppose we want this change to minimize the probability of $\bx'$ being classified as positive. With a classifier $f(\bx)$, such an $\bx'$ can be obtained by minimizing $f(\bx)$ over the features of the new version $\bx'$.

However, for some physical or economical reasons, we cannot search for the optimal $\bx$ over the whole feature space $\mathbb{R}^p$. In particular, we assume the features $\{1,2,\dots,p\}$ can be partitioned into two subsets $C$ and $U$. Given a feature vector $\bx$,  let $\bx_C$ and $\bx_U$ represent the sub-vectors of $\bx$ that contain only changeable and only unchangeable features, respectively. Since $\bx_U$ cannot be changed, we will minimize $f(\bx)$ by optimizing $\bx_C$. Hence, we represent $f(\bx)$ as $f(\bx_U,\bx_C)$ to distinguish these two sub-vectors.
In addition, we assume the reasonable value of each changeable feature in $C$ must be within an interval, denoted by $[l_i,u_i]$ for $i\in C$. Moreover, the costs for increasing and decreasing any feature $x_i$ by one unit are denoted by $c_i^+$ and $c_i^-$, respectively. Give a limited budget $B$, the optimal feature design problem for a given instance $\bx$ can be formulated as follows:
\begin{align}
\label{FeatureOpt}
\min_{\bx_C'\in\mathbb{R}^{|C|}}&f(\bx_U,\bx_C')\\\nonumber
\text{s.t.}&\sum_{i\in C}c_i^+(x_i'-x_i)_++c_i^-(x_i'-x_i)_-\leq B\\\nonumber
&l_i\leq x_i'\leq u_i\text{ for }i\in C,
\end{align}
where $(x)_+=\max\{0,x\}$ and $(x)_-=\max\{0,-x\}$.

In a more general setting, some of the features in $C$ can be changed directly by the designer. We call these features the directly changeable features. However, there are features that cannot be changed directly. Instead, they change as a consequence of manipulations made to the directly changeable features. We call these indirectly changeable features. In Chi et~al.~\cite{Chi2012} the effects of the directly changeable on the indirectly changeable features are measured upon completion of the inverse classification process. Our method incorporates them as part of the optimization.

To model this phenomenon, we further partition the features in $C$ into two subsets, $D$ and $I$, which represent the sets of directly and indirectly changeable features, respectively. When we optimize the features, we can only determine the value for $\bx_D$ and the values of $\bx_I$ will depend on $\bx_D$ and $\bx_U$. Therefore, we model the dependency of $\bx_I$ on $\bx_D$ and $\bx_U$ as $\bx_I=H(\bx_D,\bx_U)$ where the mapping $H:\mathbb{R}^{|D|+|U|}\rightarrow\mathbb{R}^{|I|}$ is assumed to be smooth and differentiable. Note that the mapping $H$ can be trained using the same training instances for $f(\bx)$. Furthermore, while the estimates elicited from $H$ may be noisy, using $H$ is better than allowing the $I$ values to remain static by definition of what $I$ represents. Therefore, we represent $f(\bx)$ as $f(\bx_U,\bx_I,\bx_D)$ to distinguish these three blocks so that the feature optimization problem \eqref{FeatureOpt} can be generalized to
\begin{align}
\label{FeatureOptGen}
\min_{\bx_D'\in\mathbb{R}^{|D|}}&f(\bx_U,H(\bx_{D}',\bx_U),\bx_D')\\\nonumber
\text{s.t.}&\sum_{i\in D}c_i^+(x_i'-x_i)_++c_i^-(x_i'-x_i)_-\leq B\\\nonumber
&l_i\leq x_i'\leq u_i\text{ for }i\in D.
\end{align}

We relate a specific method for solving $H(\bx_{D}',\bx_U)$ in Section IV.A.3. We note that, in practice, $D$ is likely to be small and that, while $U$ may be large (e.g.,~pictorial or text-based features), the efficiency of the optimization won't be affected.

\textcolor{black}{
\subsubsection{Time Complexity of $H$}
We acknowledge that the size of the indirectly changeable feature set $I$ may be large and, as a result, wish to examine the time complexity associated with the indirect feature estimator $H$, which may prove to be a computational bottleneck.}

\textcolor{black}{Let $H_a$ denote the indirect feature estimator for feature $a \in I$ and let $r_a$ denote the corresponding time complexity associated with using $H_a$; that is, $H_a$ is $\mathcal{O}(r_a)$. We can then write the time complexity of $H$ as
	\begin{align}
	R = \sum\limits_{a \in I} r_a
	\end{align}
	where $R$ is the time complexity of $H$. As we can see, $R$ increases linearly with the size of $I$ (this is by virtue of the fact that we can estimate each feature in $I$ independently). However, depending on the choice of $H_a$, and the size of $I$, this may still prove to be a bottleneck. If this is the case, the user may need to tailor their selection of $H_a$, or forgo estimating certain $I$ features during the inverse classification process. We empirically show that the time complexity scales linearly using the $H$ defined in the experiments section (kernel regression), and include the result in the supplementary material that can be found at the publicly accessible repository {\sffamily \textcolor{dgreen}{github.com/michael-lash/BCIC}}.
}

\subsubsection{Hard-line and elastic bound-setting methods}

The constraints in \eqref{FeatureOpt} and \eqref{FeatureOptGen} are flexible enough to model different feature perturbation requirements. Specifically, there are two ways that the lower and upper bounds can be parameterized, each resulting in different algorithmic behavior.

The first is rigid with respect to test point $\bx$'s original directly changeable values: if $c_{i}^{-} = 0$ then $l_{i} = x_{i}$, and if $c_{i}^{+} = 0$ then $u_{i} = x_{i}$ where $i \in D$. Such box constraint parameterization prevents feature $i$ from being increased without cost if $c_{i}^{+} = 0$, or from being decreased without cost if $c_{i}^{-} = 0$, even if doing so would be beneficial according to the local function space, determined by $f(\bx)$. This allows for more control over the recommendations being made to individuals and is most appropriate when domain experts can interject their own knowledge in designating which directions of change
are most beneficial. We refer to this as the Hardline bound-setting method.


The second is less rigid, allowing feature $i$ to increase even if $c_{i}^{+} = 0$, or to decrease even if $c_{i}^{-} = 0$. To obtain such behavior, if $c_{i}^{+} = 0$ then $u_{i} = \max\{1,x_{i}\}$ and if $c_{i}^{-} = 0$ then $l_{i} = \min\{0,x_{i}\}$. We refer to this as the Elastic bound-setting method.

In practice, we acknowledge any combination of these bound-setting methods can be used in a feature-specific manner. Bounds and costs can also be imposed such that individual costs are incurred differently, depending on whether a specific feature is increased or decreased.

\subsection{Optimization Method}

To solve the inverse classification problem, according to \eqref{FeatureOpt} and \eqref{FeatureOptGen}, we assume that objective function $f$ is differentiable and its gradient is Lipschitz continuous. Under this assumption, if $f$ is linear, the problem can be solved optimally and efficiently. If, however, the objective function is highly non-linear and non-convex, finding the globally optimal solution is NP-hard, in general. Because we do no wish to make further assumptions about the linearity of $f$, we focus on methods that can solve both these and the harder non-linear, non-convex class of function.

The available techniques that can be applied to non-convex, constrained optimization problems (see \cite{Neumaier:04} and extensive references therein) include: (a) deterministic approaches such as branch and bound~\cite{Neumaier:04}, function approximation~\cite{Jones:01}, cutting plane methods~\cite{Tuy:85}, difference of convex functions methods~\cite{Tuy:09}; and (b) stochastic approaches such as genetic algorithms~\cite{Goldberg:89}. However, these methods are typically slow and do not scale to large problems\footnote{This fact is observed first-hand in conducting our own experiments; such an experience will be further elaborated on in Section IV.}.

Therefore, our list of potential methods is left to include the projected/proximal gradient method~\cite{Nesterov07composite,Ghadimi:13a} and the zero-order method~\cite{Ghadimi:13a}. If $f(x)$ is second-order differentiable, the list of potential methods can be extended to include regularized Newton's method, sequential quadratic programming and BFGS. Among these methods, the projected gradient method and the zero-order method can guarantee that the iterative solution converge to a stationary point at a rate of $O(\frac{1}{t})$. The remaining methods only guarantee asymptotic convergence, with no specified convergence rate. Since the zero-order method is appropriate only when evaluating the gradient of $f$ is difficult, which is not our case, the appropriate method to apply with good theoretical guarantees is the projected gradient method.

\subsubsection{The Projected Gradient Method}

Before we present the projected gradient method, we need to reformulate \eqref{FeatureOpt} or \eqref{FeatureOptGen} using the difference of the original features and updated features as our decision variables. Because space is limited, we will only conduct the reformulation and presentation of the algorithm for \eqref{FeatureOptGen}, but the same technique can be applied to \eqref{FeatureOpt}. In \eqref{FeatureOptGen}, we define $\bz=\bx_D'-\bx_D$ and, by changing variables, \eqref{FeatureOptGen} can be equivalently written as
\begin{eqnarray}
\label{FeatureOptGenReform}
\min_{\bz\in\Delta_{D}}&&g(\bz)
\end{eqnarray}
where $g(\bz)\equiv f(\bx_U,H(\bx_D+\bz,\bx_U),\bx_D+\bz)$,
\begin{align}
\label{FeasibleSet}
\Delta_{D}\equiv\left\{\bz\in\mathbb{R}^{|D|}\bigg|
\begin{array}{c}
\sum_{i\in D}c_i^+(z_i)_++c_i^-(z_i)_-\leq B,\\
l_i'\leq z_i\leq u_i'\text{ for }i\in D.
\end{array}
\right\},
\end{align}
$l_i'=l_i-x_i$ and $u_i'=u_i-x_i$ for $i\in D$. The projection mapping onto the set $\Delta_{D}$ is defined as
\begin{eqnarray}
\label{proj}
\textbf{Proj}_{\Delta_{D}}(\bw)\equiv\argmin_{\bz\in\Delta_{D}}\frac{1}{2}\|\bz-\bw\|^2.
\end{eqnarray}
When $g(\bz)$ is differentiable and its gradient $\nabla g(\bz)$ is $L$-Lipschitz continuous,\footnote{ $\nabla g(\bz)$ is $L$-Lipschitz continuous if $\|\nabla g(\bz)-\nabla g(\bz')\|\leq L\|\bz-\bz'\|$ for any $\bz,\bz'\in\mathbb{R}^{|D|}$.} which is true for our class of function, the projected gradient method for solving \eqref{FeasibleSet} is then given as Algorithm \ref{algo:PGM}.

\begin{algorithm}
	\caption{Projected Gradient Method}
	\label{algo:PGM}
	\begin{algorithmic}[1]
        \REQUIRE{$\bz^{(0)}\in\Delta_{D}$, $t=0$ and $\eta>0$}
		\WHILE {Stopping criterion is not satisfied}
        \STATE {$\bz^{(t+1)}=\textbf{Proj}_{\Delta_{D}}(\bz^{(t))}-\eta\nabla g(\bz^{(t)})$}
        \STATE {$t\leftarrow t+1$}
        \ENDWHILE
        \ENSURE $\bz^{(t)}$
	\end{algorithmic}
\end{algorithm}

 According to Theorem 3 of \cite{Nesterov07composite}, when $\eta\leq\frac{1}{L}$, Algorithm~\ref{algo:PGM} guarantees that $\bz^{(t)}$ converges to a stationary point (or so-called KKT point) of \eqref{FeatureOptGenReform} at a rate of $O(\frac{1}{t})$, which is the best convergence for non-convex smooth optimization.

 Algorithm~\ref{algo:PGM} requires solving the projection $\textbf{Proj}_{\Delta_{D}}(\bw)$ at each iteration, which is itself an optimization problem. An efficient solution scheme for this subproblem is critical for making Algorithm~\ref{algo:PGM} expeditious. Fortunately, the domain $\Delta_{D}\neq\emptyset$ has a specific structure which allow us to solve $\textbf{Proj}_{\Delta_{D}}(\bw)$ for any $\bw$ with an efficient subroutine. To see this, we define
\begin{align}
\label{h}
h_i(w,\lambda)=
\left\{
\begin{array}{ll}
w-\lambda c_i^+&\text{ if }\lambda\leq \frac{w}{c_i^+}\text{ and }w> 0\\
w+\lambda c_i^-&\text{ if }\lambda\leq -\frac{w}{c_i^-}\text{ and }w< 0\\
0&\text{ otherwise}
\end{array}
\right.
\end{align}
for each $i\in D$. The subroutine is given in Algorithm \ref{algo:Proj}.
\begin{algorithm}[h]
	\caption{Projection Mapping $\textbf{Proj}_{\Delta_{D}}(\bw)$}
	\label{algo:Proj}
	\begin{algorithmic}[1]
        \REQUIRE $\bw\in\mathbb{R}^{|D|}$, $\{c_i^+\}_{i\in D}$, $\{c_i^-\}_{i\in D}$, $\{l'_i\}_{i\in D}$ and $\{u'_i\}_{i\in D}$
		\STATE {$\mathcal{A}_-\leftarrow\{i|u_i'\leq \min(0,w_i)\}$
        \STATE $\mathcal{A}_+\leftarrow\{i|\max(0,w_i)\leq l_i'\}$}
        \STATE {$z_i\leftarrow u_i'$ for $i\in\mathcal{A}_-$ and $z_i\leftarrow l_i'$ for $i\in\mathcal{A}_+$}
        \IF {\small$\sum_{i\in D\backslash(\mathcal{A}_+\cup \mathcal{A}_-)}\max\{\min\{h_i(w_i,0),u_i'\},l_i'\}
        \leq B-\sum_{i\in\mathcal{A}_-}u_i'c_i^--\sum_{i\in\mathcal{A}_+}l_i'c_i^+$\normalsize}
        \STATE{$\lambda\leftarrow 0$}
        \ELSE
        \STATE{Apply bisection search to find $\lambda\in(0,+\infty)$ such that
        \small
        \begin{eqnarray*}
        &&\sum_{i\in D\backslash(\mathcal{A}_+\cup \mathcal{A}_-)}\max\{\min\{h_i(w_i,\lambda),u_i'\},l_i'\}\\
        &=& B-\sum_{i\in\mathcal{A}_-}u_i'c_i^--\sum_{i\in\mathcal{A}_+}l_i'c_i^+
        \end{eqnarray*}
        \normalsize}
        \ENDIF
        \STATE {$z_i\leftarrow\max\{\min\{h_i(w_i,\lambda),u_i'\},l_i'\}$ for $i\in D\backslash(\mathcal{A}_+\cup \mathcal{A}_-)$}
        \ENSURE $\bz$
	\end{algorithmic}
\end{algorithm}

The correctness of Algorithm~\ref{algo:Proj} is ensured by the following proposition whose proof is given in the Appendix.
\begin{proposition}
\label{prop:proj}
If $\Delta_{D}\neq\emptyset$, the solution $\bz$ returned by Algorithm~\ref{algo:Proj} satisfies $\bz=\textbf{Proj}_{\Delta_{D}}(\bw)$.
\end{proposition}

\textcolor{black}{
\subsection{Representativeness and Support}
}

\textcolor{black}{With our methodology defined, we wish to comment on, and subsequently quantify, both the representativeness of the training set from which our $f$ will generalize and the support underlying the inverse classification of an instance.}
%
\textcolor{black}{
Therefore, we first propose \textbf{$\pmb{\delta}$-dissimilarity}, related by Definition \ref{def:drep}, which quantifies the dissimilarity between the training set distribution $S$ and population distribution $\mathcal{P}$ using a linear discrepancy distance measure defined in Johansson et~al.~\cite{Johansson2016}.
\begin{definition}
	\label{def:drep}
	The distribution $S$ of the training set, drawn from the population distribution $\mathcal{P}$, is said to be \textbf{$\pmb{\delta}$-dissimilar} to that of $\mathcal{P}$ if
	\begin{align}
	\text{disc}_{\mathcal{H}}(S,\mathcal{P}) \leq \delta.
	\end{align}
	where $\text{disc}_{\mathcal{H}}(S,\mathcal{P}) \triangleq \lVert \mu(S) - \mu(\mathcal{P})\rVert$ is the \textit{discrepancy distance} between two samples \cite{Johansson2016}, or in this case the training sample and population, we define $\mu(\cdot)$ to denote the mean of a particular distribution, and $\lVert \cdot\rVert$ is the Euclidean norm.
\end{definition}
Using Definition \ref{def:drep}, we relate the following proposition.
\begin{proposition}
	\label{prop:0rep}
	 As the size of the training set $n$ increases to infinity, the training set distribution $S$ is asymptotically $\delta=0$-dissimilar to that of population distribution $\mathcal{P}$.
\end{proposition}
The proof of Proposition \ref{prop:0rep} is in the appendix. We wish to point out, however, that the variance and shape of $\mathcal{P}$ and $S$ may be quite different despite $S$ being $\delta=0$-dissimilar to that of $\mathcal{P}$. Additionally, in practice, the i.i.d. assumption may not hold (in this work we assume it does). We leave methods, taking into account such factors, as tangential future work.}

\textcolor{black}{
We are also concerned with ensuring that optimized instances be near training data. These underlying training data provide support as to the ``trustworthyness'' of the recommendations and corresponding probabilities elicited from the inverse classification process. Therefore, we define \textbf{$\pmb{(\epsilon,\gamma)}$-support}, related by Definition \ref{def:egrep}, which empirically quantifies the degree to which an inversely classified instance can be trusted.
\begin{definition}
	\label{def:egrep}
	Define the \textbf{$\pmb{(\epsilon,\gamma)}$-support} for a particular test instance $\bx$, to be the following:
	\begin{itemize}
		\item[-] $\epsilon$ is the variance in the predicted probabilities of $\bx$'s $k$ nearest neighbors (from the training data). This measure provides an assessment as to the stability of the local probability space surrounding $\bx$.
		\item[-] $\gamma$ is the number of neighbors that fall within $\bar{\text{maxDkNN}}=\frac{1}{n}\sum_{i=1}^{n}\text{maxDist}(k\text{NN}(\bx^i))$ of $\bx$, where the function $\text{maxDist}(\cdot)$ returns the maximum distance of training instance $\bx^i$'s $k$ nearest neighbors; $\bar{\text{maxDkNN}}$ represents the average of these maximum distances. By comparing the $\gamma$ of $\bx$ to the average $\gamma$ of the training set we can observe whether a particular test instance has more (larger $\gamma$) or less (smaller $\gamma$) ``support'' (relative to the average from the training data) underlying the predicted probability.
	\end{itemize}
\end{definition}
We explore $(\epsilon,\gamma)$-support in the Experiments section.}

\begin{figure}[b]
	\centering
	\includegraphics[scale=.34]{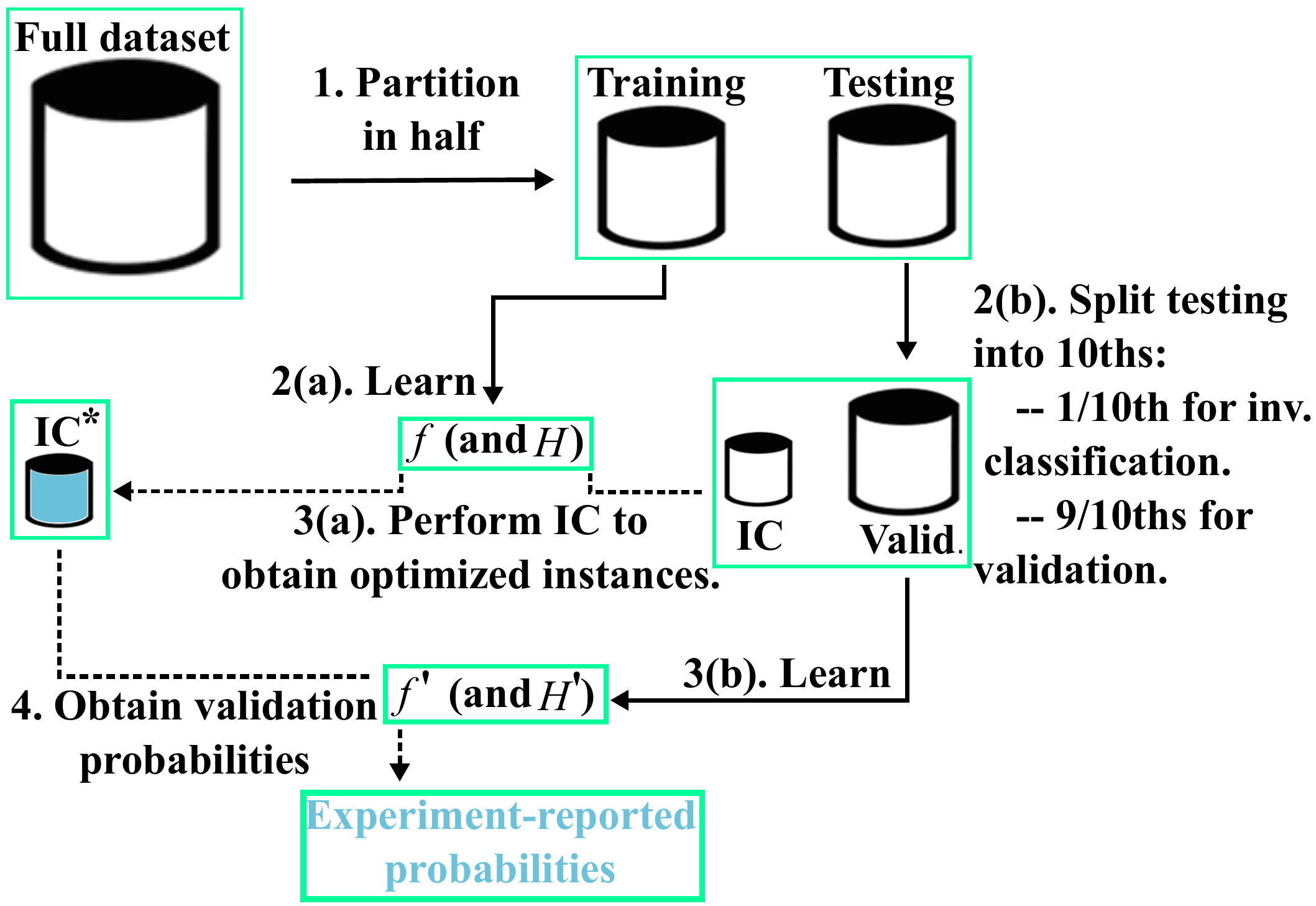}
	\caption{Experiment process.\label{fig:exp}}
\end{figure}

\section{Experiments}

In this section we outline our experimental methods and then apply such methods to two datasets. The first is a benchmark dataset from the UCI Machine Learning Repository \cite{uci} called Student Performance \cite{cortez2008}. The second is derived from ARIC, the Atherosclerosis Risk in Communities study \cite{aric1989}. We emphasize that both datasets are publicly available. The latter requires explicit NIH permission\footnote{Obtained via BioLINCC.}. We provide the code used in all experiments, and processed Student Performance data for public use at {\sffamily \textcolor{dgreen}{github.com/michael-lash/BCIC}}. The list of unchangeable, indirectly changeable, and directly changeable features (and corresponding parameters) for both datasets is also provided at the above mentioned URL.

\textcolor{black}{We emphasize that parameterization of the inverse classification framework, including the costs-to-change and assignment of features to the categories of unchangeable, indirectly changeable and directly changeable, should be guided by domain experts. As such, our experiments on the ARIC dataset are guided by a CVD specialist who is a co-author of this work.}

\subsection{Experiment Parameters and Setup}

In this section we outline a general process of validating inverse classification methods, the two learning algorithms used to conduct the inverse classification, a method for estimating indirectly changeable features, and a benchmark optimization method which we will compare against our gradient-based method.

\subsubsection{Process}

Our process of making and evaluating recommendations is based on that proposed by \cite{Chi2012}. In our experiments, we are using data from the past in which known outcomes are observed. We then make recommendations that reduce the probability of a negative outcome occurring. But, in the absence of a time machine, we need a way of validating whether we would have actually reduced the probability of such an event occurring. A method that accomplishes this requires careful segmentation of the data such that none of the information used to make recommendations is used in validating the probability of an outcome occurring. The process, shown in Figure \ref{fig:exp}, is related as follows:



\textbf{Step 1: }Partition the full dataset into two equal parts: a training set and a testing set. Data cleansing and preparation are also performed, including missing value imputation (mean) and the normalization of data values to be within $[0,1]$.

\textbf{Step 2(a): }uses the training set to learn a model $f$. During this step cross-validation can be used to find the optimal parameters of $f$, if necessary. We also perform cross-validation to obtain optimal parameters in the model $x_I=H(x_D,x_U)$ for indirectly changeable features.

\textbf{Step 2(b): }Further split the testing set into 10ths. 1/10th is for performing inverse classification on and the other 9/10ths are used for validation.

\textbf{Step 3(a): } Perform inverse classification on the heldout 10th of data using $f$.

\textbf{Step 3(b): } Learn a validation $f^{\prime}$ (and $H^{\prime}$) using the 9/10ths of heldout testing data.

\textbf{Step 4: } Estimate probabilities for the optimized inverse classification instances using $f^{\prime}$. These are the probabilities we report in our experiments. Note that we obtain probabilities for each 1/10th of held out testing data.

\textcolor{black}{By setting up the experiment in this manner we are also able to be more confident that the recommendations obtained are not the result of overfitting. Note also that by switching the roles of training and validation/test sets, the full amount of data can be used to obtain results.}

\subsubsection{Classification Functions}

Our experiments employ the use of two different learning methods: the linear \textit{logistic regression} model and the nonlinear \textit{kernel SVM}.

Logistic regression is a popular predictive model that works particularly well when the linear feature independence assumption holds. The model is trained via maximum likelihood estimation, given by the optimization problem

\begin{align}
\label{eq:logregopt}
\max\limits_{\bb, \beta_{0}} \sum\limits_{i=1}^{n} -\text{log}( 1 + \text{exp}(\beta_{0}+\bb^{\top}\bx^{i})) + \sum\limits_{i=1}^{n} y^{i}(\beta_{0}+ \bb^{\top}\bx^{i})
\end{align}
where $\bb$ and $\beta_{0}$ are a vector of coefficients and offset term, respectively. After being trained the $\bb$ and $\beta_{0}$ can be used to make classifications for a given test instance $\bx$ by
\begin{align}
\label{eq:logclass}
f(\bx) = \frac{1}{1 + \text{exp}(-(\beta_{0} + \bb^{\top}\bx))}
\end{align}
which gives the probability of $\bx$ being in the positive class.

Employment of the logistic model in our described inverse classification framework can be viewed as a basic method having roots in sensitivity analysis. This is illustrated by observing the link between coefficient examination as a means of sensitivity analysis and the employment of our described gradient-based methodology. Examining the sign and magnitude of a coefficient uncovers a particular feature's bearing -- how positive or how negative -- on the problem being modeled. Taking the gradient of a linear model has the same effect, thus informing the inverse classification framework which feature perturbations decrease the objective function value, with larger coefficients having a larger effect. Integration of this optimization methodology into the framework allows cost, budget, etc.~to be taken into account as well.

Among classification models, the kernel SVM is one of the most widely used. Compared to the classical linear SVM, kernel SVM is more appropriate for data in which two classes of instances have a nonlinear boundary.
A kernel SVM model can be trained using its dual formulation which is related by the optimization problem
\begin{align}
\label{SVMdual}
\max_{\ba\in\mathbb{R}^n}&\sum_{i=1}^n\alpha_i-\frac{1}{2}\sum_{i=1}^n\sum_{j=1}^n\alpha_i\alpha_jy^iy^jk(\bx^i,\bx^j)\\\nonumber
\text{s.t.}&\sum_{i=1}^n\alpha_iy^i=0\text{ and }0\leq \alpha_i\leq C\text{ for }i=1,2\dots,n,
\end{align}
where $k(\bx,\bx'):\mathbb{R}^p\times \mathbb{R}^p\rightarrow \mathbb{R}$ is a kernel function that measures the similarity between any pair of instances $x$ and $x'$ in $\mathbb{R}^p$. The commonly used kernel functions include linear kernels $k(\bx,\bx')=\bx^T\bx'$, polynomial kernels $k(\bx,\bx')=(1+x^Tx')^d$ for any positive integer $d$, and Gaussian kernels $k(\bx,\bx')=\exp\left(-\frac{\|\bx-\bx'\|^2}{2\sigma^2}\right)$ for $\sigma>0$ where $\|\cdot\|$ represents the Euclidean norm in $\mathbb{R}^p$.

Suppose the optimal solution of \eqref{SVMdual} is $\ba^*\in\mathbb{R}^n$. An SVM classifier can be derived based on the function\footnote{In fact, the exact kernel SVM classifier is $f_b(\bx)=\sum_{i=1}^n\alpha_i^*y^ik(\bx^i,\bx)+b$ where $b$ is an offset value such that the new instance $\bx$ is classified to be positive if $f_b(\bx)>0$ and to be negative otherwise. }
\begin{eqnarray}
\label{simfun}
f(\bx)=\sum_{i=1}^n\alpha_i^*y^ik(\bx^i,\bx),
\end{eqnarray}
where the instance $\bx^i$ with $\alpha_i^*>0$ is called a support vector.
Given a new instance $\bx$, the value of $f(\bx)$ represents how similar $\bx$ is to the positive class. A larger value of $f(\bx)$ means that $\bx$ is more likely to be positive.

However, the scores obtained from $f(\bx)$ do not correspond to likelihood directly. Therefore, we apply Platt's Method \cite{platt1999}. Platt's Method transforms the scores obtained from applying $f(\bx)$ to probabilities; specifically, the probability of being positive. By applying this method we learn a probability space that is more easily interpretable.

We elect to use the Gaussian kernel SVM for three reasons. The first is that such a function is highly nonlinear and complex, giving us the opportunity to explore a more flexible classifier by which we can assess the effectiveness of our method. Secondly, the Gaussian kernel can be used to assess point similarity. This is beneficial in our experiments as one of our assumptions is that similar points will have similar probabilities associated with them, which isn't enforced by linear predictors. Finally, using the $\sigma$ parameter, we can control the size of the neighborhood used to assess point similarity. That is, larger $\sigma$ values make more distant support vectors appear more similar to a test point $\bx$, which subsequently has the effect of smoother probability transitions during optimization.

Therefore, our objective function, outlined in (\ref{FeatureOpt}) and (\ref{FeatureOptGen}), becomes \eqref{eq:logclass} and (\ref{simfun}), logistic and SVM, respectively, with features segmented into appropriate groups and the indirect feature estimator, outlined in the next subsection, incorporated. We explicitly note that, in the case of \eqref{simfun}, the minimization task is to minimize the SVM score. More appropriately, by applying Platt's method, we will be minimizing probability directly, as we are when using \eqref{eq:logclass}.

\subsubsection{Estimating Indirectly Changeable Features}

We employ the use of Kernel Regression \cite{Nadaraya1964,Watson1964} as a means of estimating the indirectly changeable features. In particular, the model $\bx_I=H(\bx_D,\bx_U)$ used in \eqref{FeatureOptGen} is
\begin{eqnarray}
\label{kernelreg}
\bx_I &=& \frac{\sum_{i=1}^{n} k([\bx^i_D, \bx^i_U],[\bx_D, \bx_U])\bx_I^i}{\sum_{i=1}^{n} k([\bx^i_D, \bx^i_U],[\bx_D, \bx_U])},
\end{eqnarray}
where the kernel $k(\bx,\bx')=\exp\left(-\frac{\|\bx-\bx'\|^2}{2\sigma^2}\right)$ (Gaussian) and the value $\sigma > 0$ is selected based on cross-validation. By using the model in \eqref{kernelreg} with the Gaussian kernel we are provided with the added benefit of a point similarity assessment in making estimations. The model works by considering the known training set $\bx_I^i$, that are closer to $\bx$, more favorably than those that are further away. In so doing, (\ref{kernelreg}) obtains an estimate for $\bx_I$ based on points that are most similar to it.

\subsubsection{Methodological Benchmark}

\textcolor{black}{In our experiments we wish to compare our method to that of another.  However, to the best of our knowledge, there exists no past methods, including those found in Section II, that can be incorporated into our framework. Therefore we develop a method, based on sensitivity analysis, that we believe represents a reasonable initial attempt at solving the problem from such a standpoint. Our proposed benchmark method operates by iteratively perturbing each feature $\bx_{D_i}$ $i \in D$ to the bounds of feasibility (and is therefore akin to the variable perturbation method of sensitivity analysis \cite{Yao2003})}. The objective function is then evaluated. If this value is found to be better than any of the previous single-feature perturbations, the perturbation is accepted. After making single-feature perturbations, if some amount of budget $B$ remains, then subsequent rounds of perturbation occur (double-feature perturbation, triple-feature perturbation, etc.).

\begin{figure*}[h]
	\centering
	\begin{subfigure}[h]{.49\linewidth}
		\centering
		\includegraphics[scale=.145]{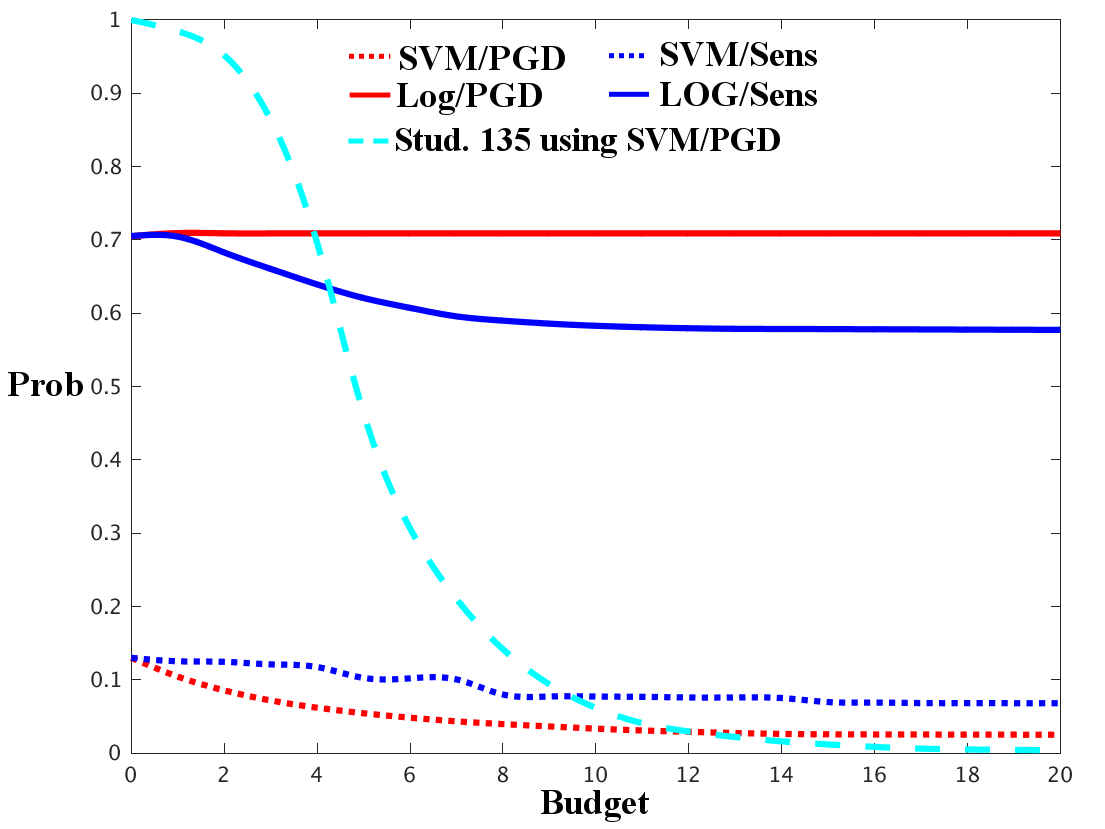}
		\caption{SP dataset using Hardline Bound-setting. \label{fig:benchS1}}
	\end{subfigure}
	\begin{subfigure}[h]{.49\linewidth}
		\centering
		\includegraphics[scale=.15]{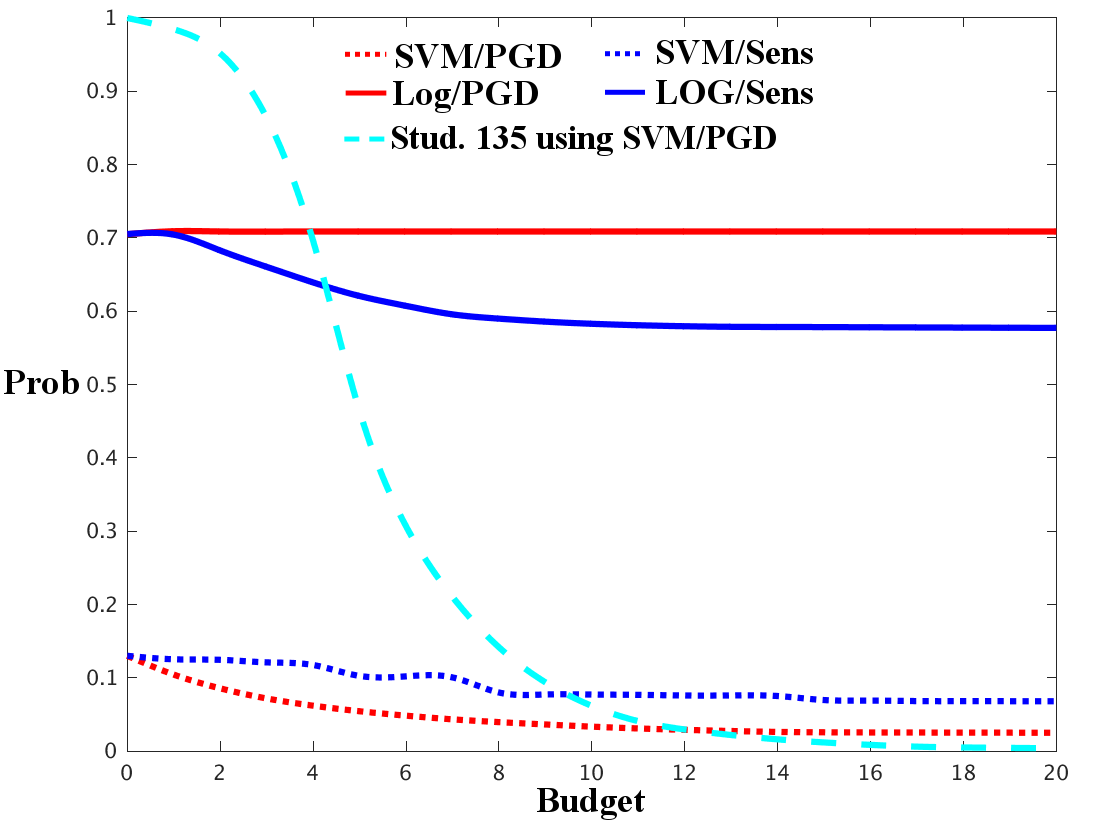}
		\caption{SP dataset using Elastic Bound-setting.\label{fig:benchS2}}
	\end{subfigure}
	\par
	\begin{subfigure}[h]{.49\linewidth}
		\centering
		\includegraphics[scale=.145]{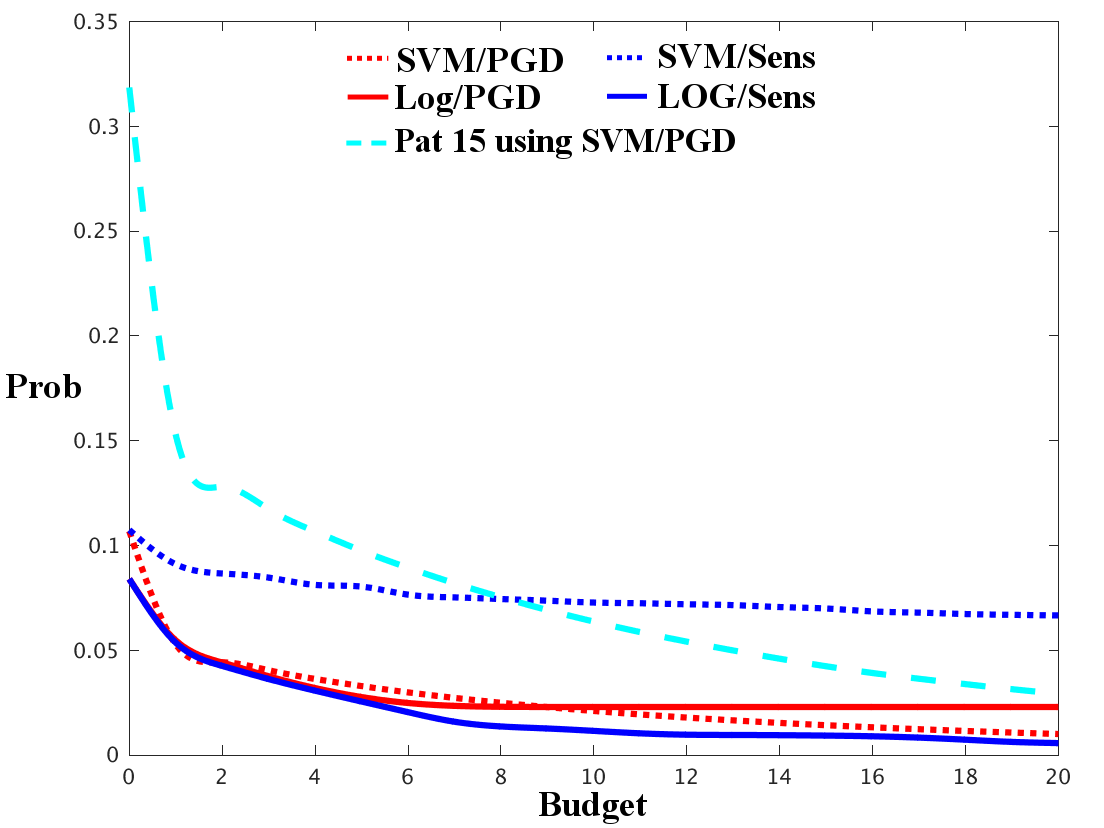}
		\caption{ARIC dataset using Hardline Bound-setting.\label{fig:aricS1}}
	\end{subfigure}
	\begin{subfigure}[h]{.49\linewidth}
		\centering
		\includegraphics[scale=.15]{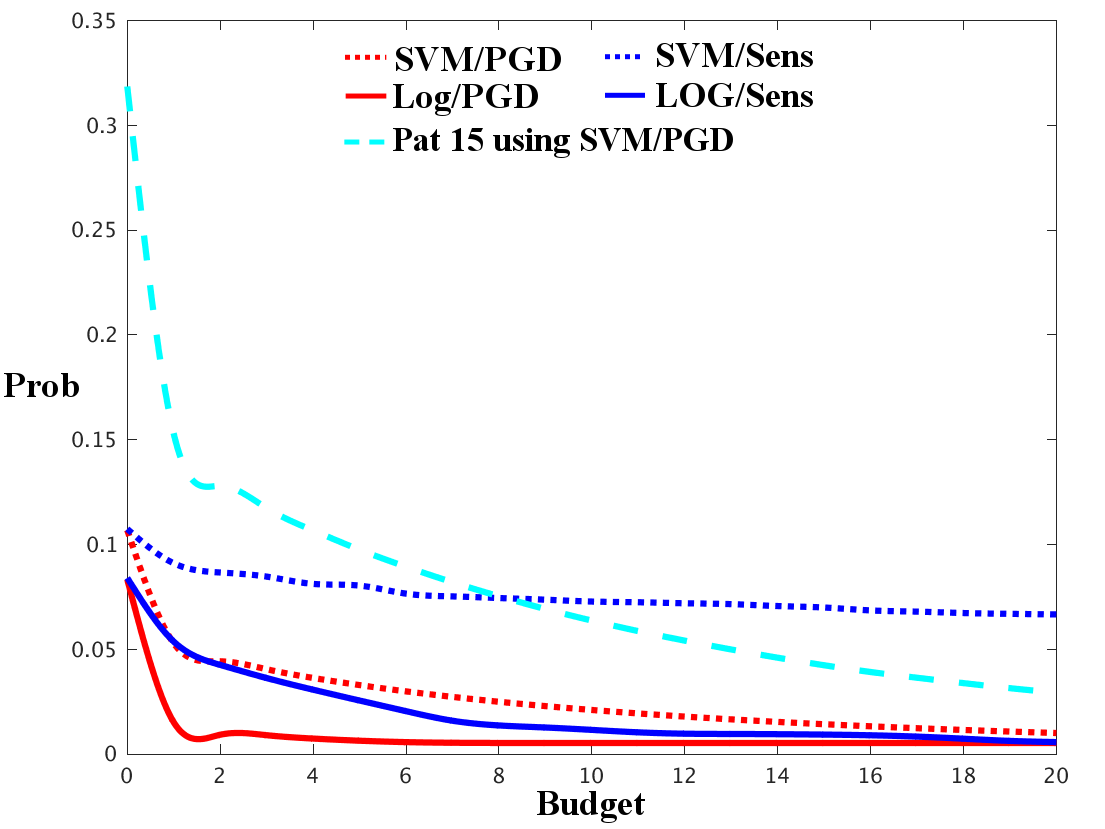}
		\caption{ARIC dataset using Elastic Bound-setting.\label{fig:aricS2}}
	\end{subfigure}
	\caption{Average probability vs.~budget by dataset (Student Performance or ARIC) and by bound-setting method. Solid lines represent a result obtained using the logistic model, while dotted lines represent a result obtained using the SVM model. PGD denotes use of the gradient method, while Sens denotes use of the sensitivity analysis-based method. The \textcolor{cyan}{cyan} dashed line is a randomly selected individual whose recommendations will be shown and discussed in the next subsection.\label{fig:avgprobres}}
\end{figure*}

\textcolor{black}{Here we assert that, because we have chosen two different indirectly changeable feature estimators, we will effectively be using two different benchmark methods.}

Cumulatively, our experiments will involve two datasets (ARIC, Student Performance), two classification functions (logistic, SVM), two optimization methods (PGD, sensitvity analysis-based), and two bound-setting methods (Hardline and Elastic) which constitute a total of 16 experiments.

\subsection{Data Description}

We validate the effectiveness of our inverse classification framework on two datasets: Student Performance and ARIC. Student Performance data consists of individual Portuguese students enrolled in two different classes. The one used in this experiment was the Portuguese language class, as it contained the greater number of instances ($n=649$). Each student-instance has 43 associated features ($p=43$). The dependent variable is whether a student earned a final grade of C or below ($y=1$) or not ($y=-1$). We discard the two intermediary grade reports to reflect the long-term goal of earning a better grade. Therefore, the task is to minimize the probability of earning a C or below.

The ARIC dataset contains $n=12907$ patients for which we define 110 features (please refer to {\sffamily \textcolor{dgreen}{github.com/michael-lash/BCIC}}). As the problem domain is medicine-based, we consulted an epidemiologist, a coauthor of this paper. We define $y=1$ to be a positive CVD diagnosis, which includes probable myocardial infarction (MI), definite MI, suspect MI, definite fatal coronary heart disease (CHD), possible fatal CHD, and stroke. Patients not having any of these diagnoses have their CVD class variable encoded as $y=-1$. Additionally, patients having one of these diagnoses prior to the study period were excluded from our dataset (giving us the final $n=12907$ patients).

\subsection{Results: Probability Reduction}

The results of our 16 experiments are reported in terms of average probability relative to budget, which can be viewed in Figure \ref{fig:avgprobres}, where the subfigures stratify results by dataset and bound-setting method.

Comprehensively we can see that, in the general case, all methods except the logistic classifier using PGD on the Student Performance dataset were successful in reducing the average probability of a negative outcome. Depending on the dataset and bound-setting method used, different methods coupled with different classifiers experienced different degrees of success. This seems to suggest that, as in typical classification settings, methodological success varies on a dataset-to-dataset basis.

Interestingly, at a high level, there is no difference between the results obtained using the Hardline and Elastic bound-setting methods on Student Performance and only one distinct difference between the results obtained on ARIC. Here, logistic regressing using the PGD method is observed to have distinctly greater average performance using the Elastic bound-setting method (shown in Figure \ref{fig:aricS2}). Such a result should be viewed cautiously, however, as the recommendations obtained may differ, and perhaps even contradict, those our cardiovascular disease specialist would view as being truly beneficial. Differences of this nature may be attributable to possible noise in the ARIC data.


In examining the results obtain on Student Performance, shown in Figures \ref{fig:benchS1} and \ref{fig:benchS2}, some interesting findings emerge \footnote{We wish to point out that the probabilistic estimates obtained from the two classifiers are disparate, which we believe stems from small amounts of training data}. We can see that the best result obtained using the logistic classifier was through the sensitivity analysis-based method and the best obtained using the SVM classifier was through PGD. This may suggest that simpler, linear classifiers may experience better inverse classification results using simpler means of optimization and that more complicated, non-linear classifiers may see better results using those that are more complicated.

\begin{figure*}[t]
	\centering
	\begin{subfigure}[]{.49\linewidth}
		\centering
		\includegraphics[scale=.165]{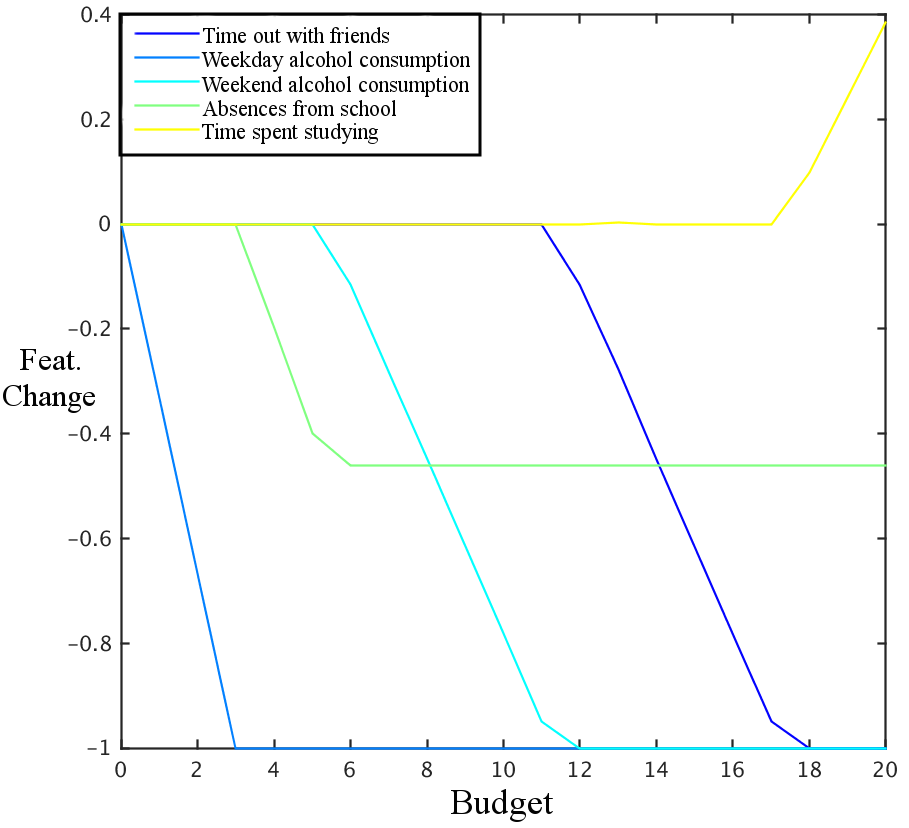}
		\caption{Student 135.\label{fig:bench_kernel_rec}}
	\end{subfigure}
	\begin{subfigure}[]{.49\linewidth}
		\centering
		\includegraphics[scale=.35]{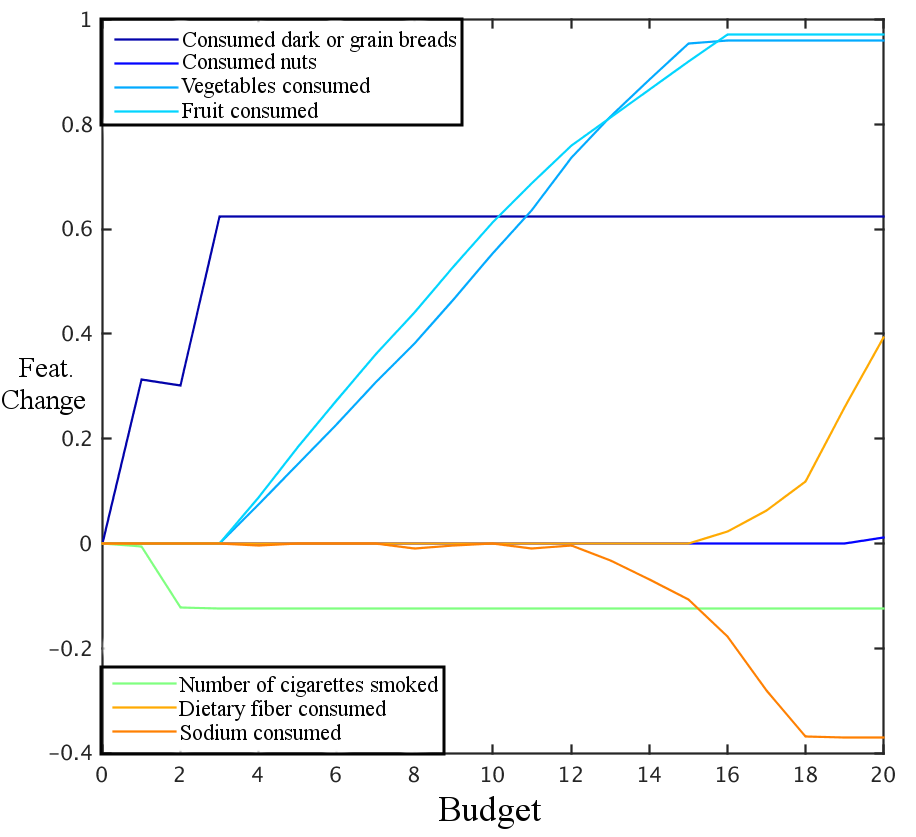}
		\caption{Patient 15. \label{fig:cvd_changes_hl}}
	\end{subfigure}
	\caption{Recommended changes vs.~budget for a randomly selected individual from each dataset. \label{fig:recommend}}
\end{figure*}

This latter point is somewhat supported by the results obtained on the ARIC dataset, shown in Figures \ref{fig:aricS1} and \ref{fig:aricS2}. In examining Figure \ref{fig:aricS1} we can see that PGD outperformed the sensitivity analysis-based method when using the nonlinear SVM classifier and that the sensitivity analysis-based method outperformed PGD when using the linear logistic classifier. However, in Figure \ref{fig:aricS2}, which represents results obtained using the Elastic bound-setting method PGD has dominated in the case of both classifiers. This result seems to suggest that, regardless of classifier complexity, if there exist optimizations that benefit from an Elastic setting (recall that no benefits were found from such a setting on Student Performance), PGD may dominate (on average).

Unexpectedly, looking at the results obtained for a randomly selected individual from either dataset, we can see that there is no difference in probabilistic improvement between the two bound-setting methods based when using SVM with PGD. The specific recommendations made to these individuals are discussed in the next subsection along with recommendations most commonly made to individuals in each dataset at a budget of four.

\begin{figure*}[t]
	\centering
	\begin{subfigure}[]{.24\linewidth}
		\centering
		\includegraphics[scale=.09]{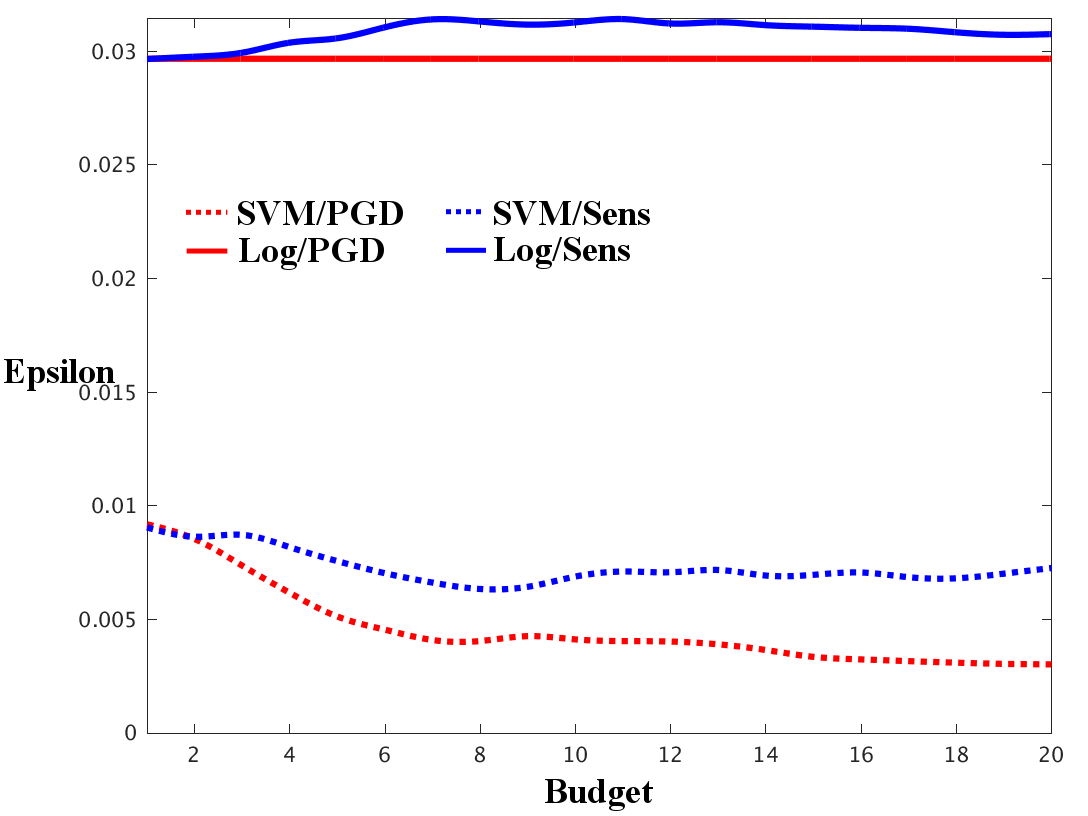}
		\caption{Stud.~Perf.~Average $\epsilon$ by budget.\label{fig:ben_eps_hl}}
	\end{subfigure}
	\begin{subfigure}[]{.24\linewidth}
		\centering
		\includegraphics[scale=.09]{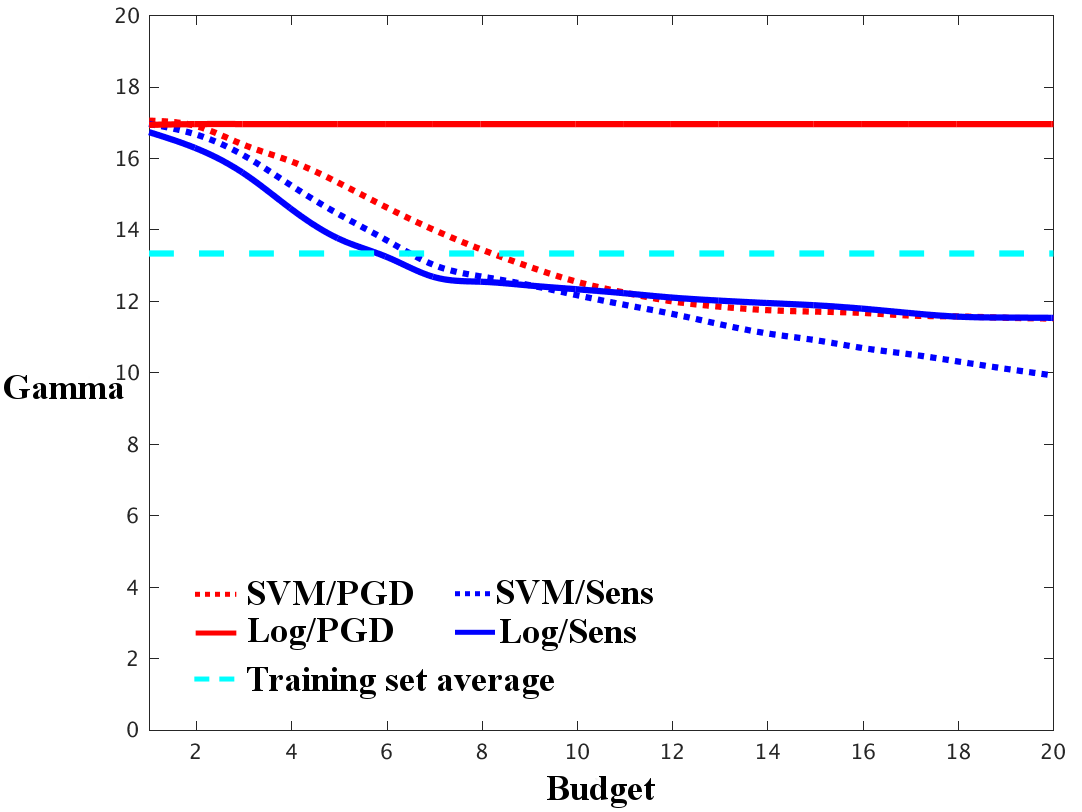}
		\caption{Stud.~Perf.~Average $\gamma$ by budget. \label{fig:ben_gam_hl}}
	\end{subfigure}
	\begin{subfigure}[]{.24\linewidth}
		\centering
		\includegraphics[scale=.09]{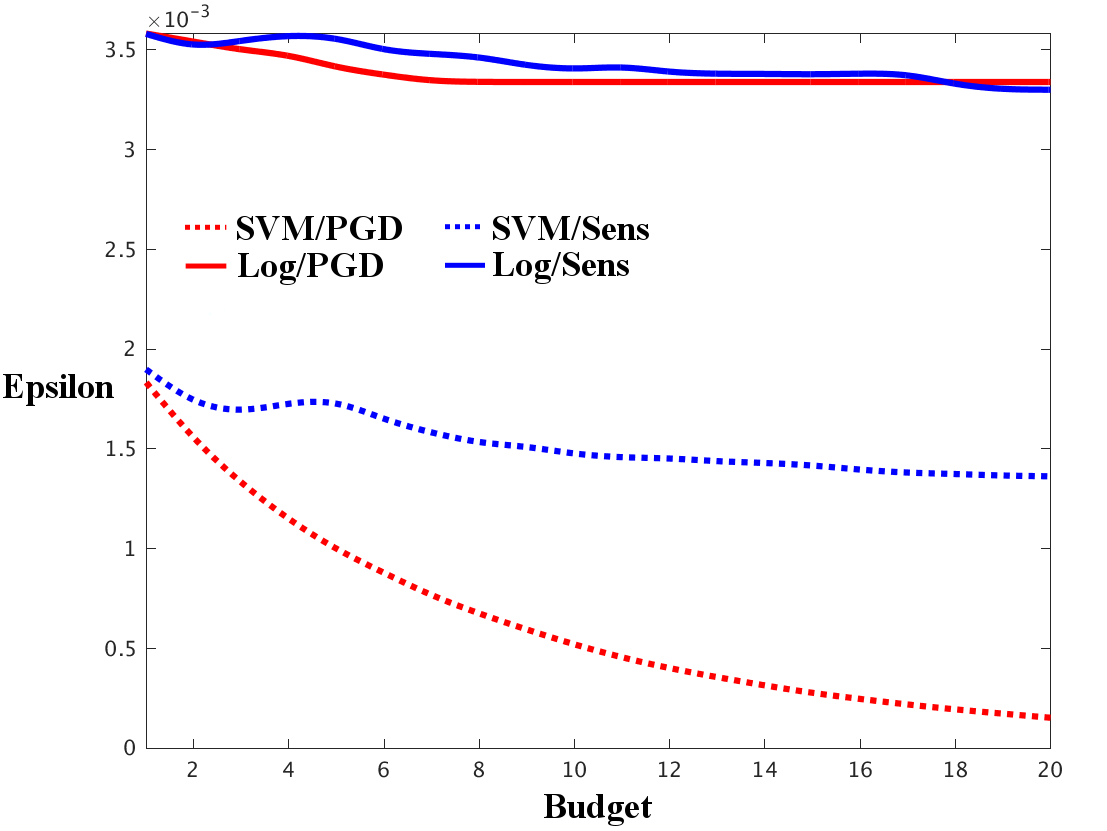}
		\caption{ARIC.~Average $\epsilon$ by budget.\label{fig:ar_eps_hl}}
	\end{subfigure}
	\begin{subfigure}[]{.24\linewidth}
		\centering
		\includegraphics[scale=.092]{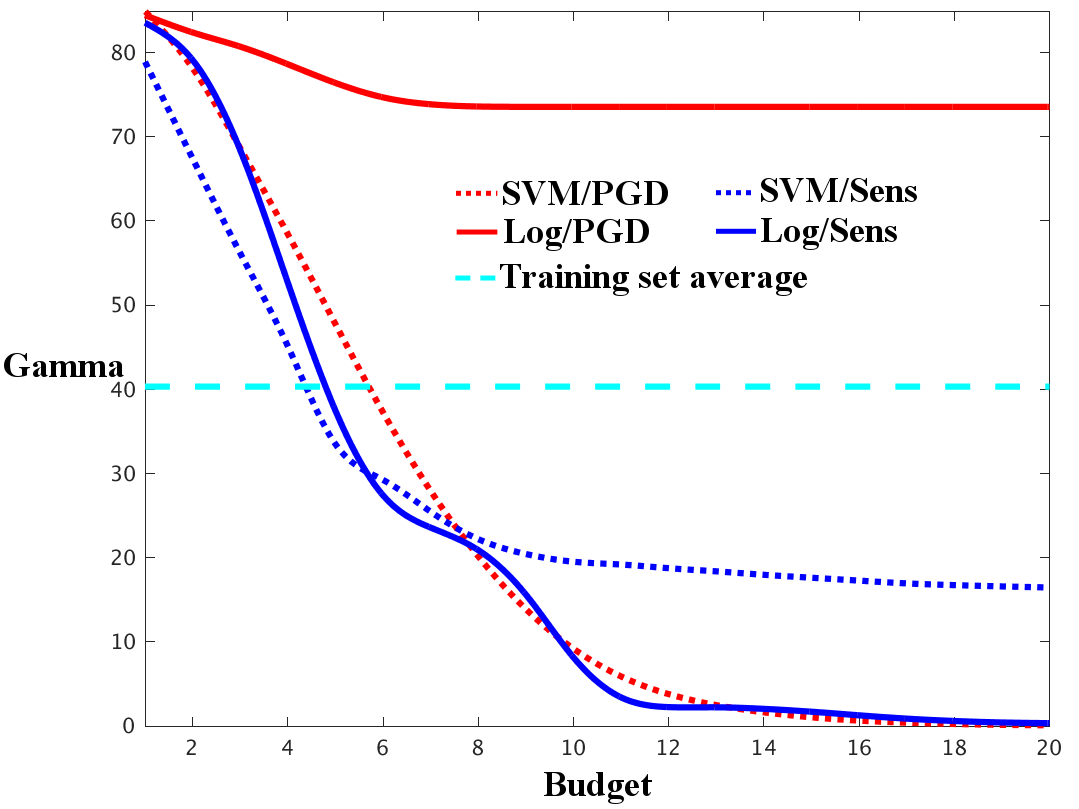}
		\caption{ARIC.~Average $\gamma$ by budget. \label{fig:ar_gam_hl}}
	\end{subfigure}
	%
	\par
	\begin{subfigure}[]{.24\linewidth}
		\centering
		\includegraphics[scale=.09]{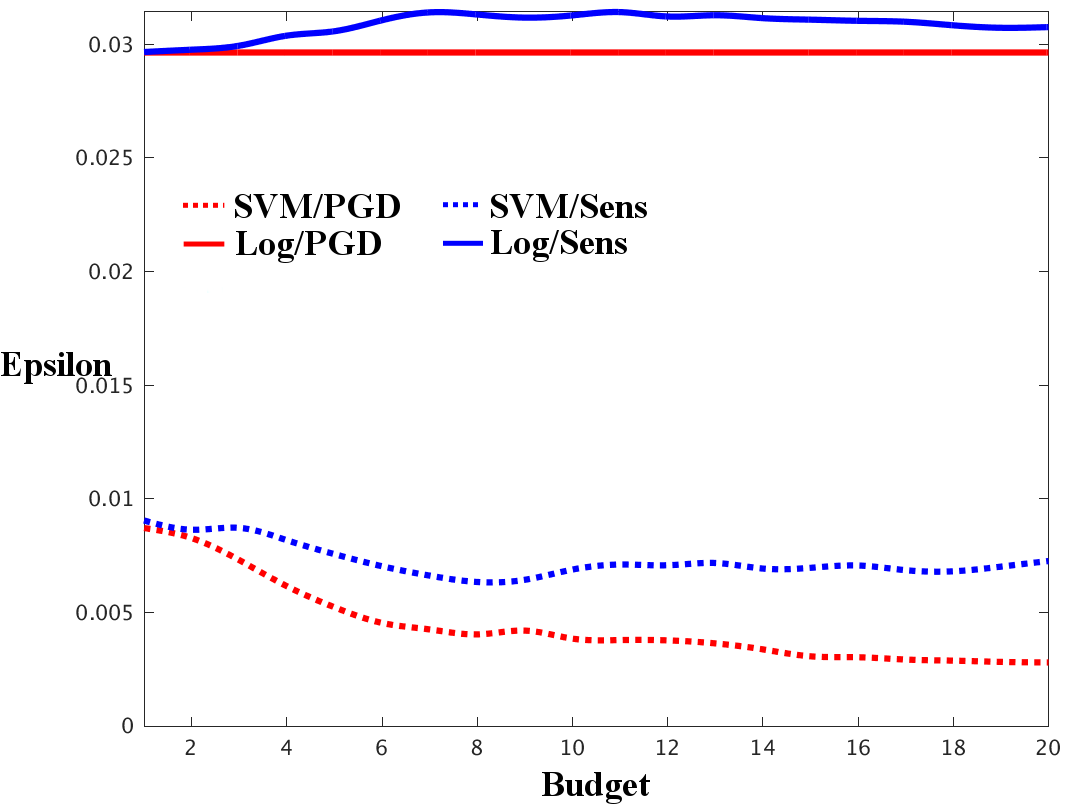}
		\caption{Stud.~Perf.~Average $\epsilon$ by budget.\label{fig:ben_eps_el}}
	\end{subfigure}
	\begin{subfigure}[]{.24\linewidth}
		\centering
		\includegraphics[scale=.09]{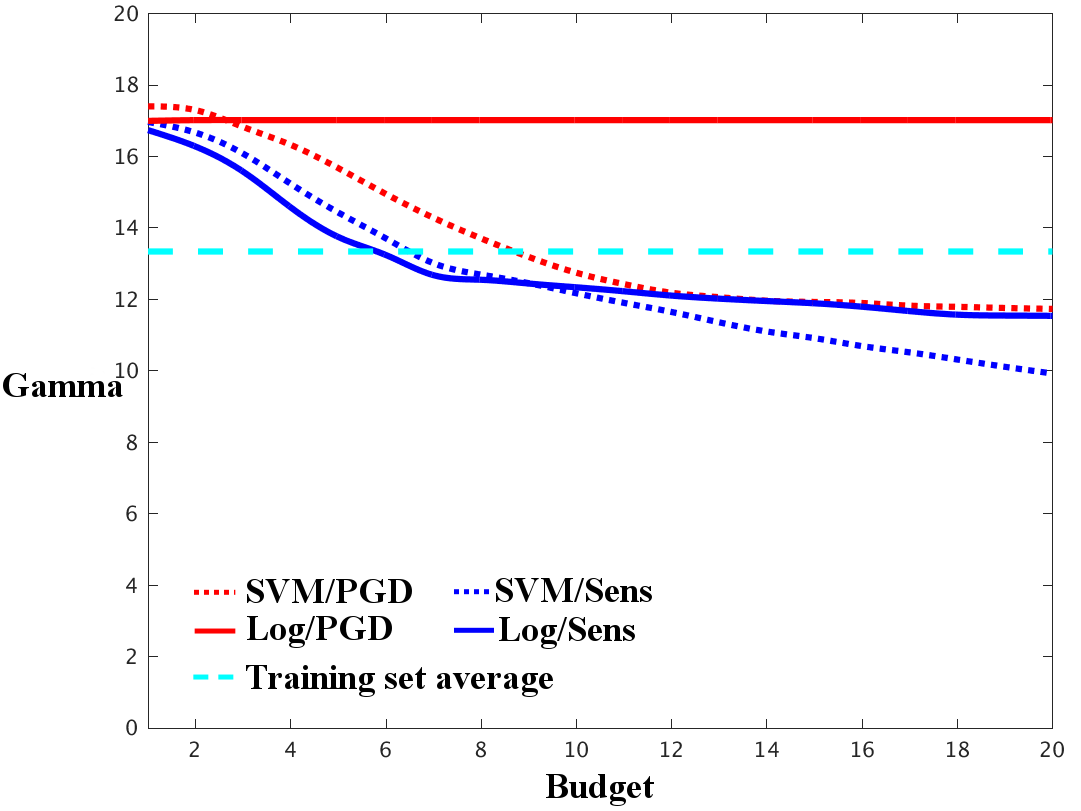}
		\caption{Stud.~Perf.~Average $\gamma$ by budget. \label{fig:ben_gam_el}}
	\end{subfigure}
	\begin{subfigure}[]{.24\linewidth}
		\centering
		\includegraphics[scale=.09]{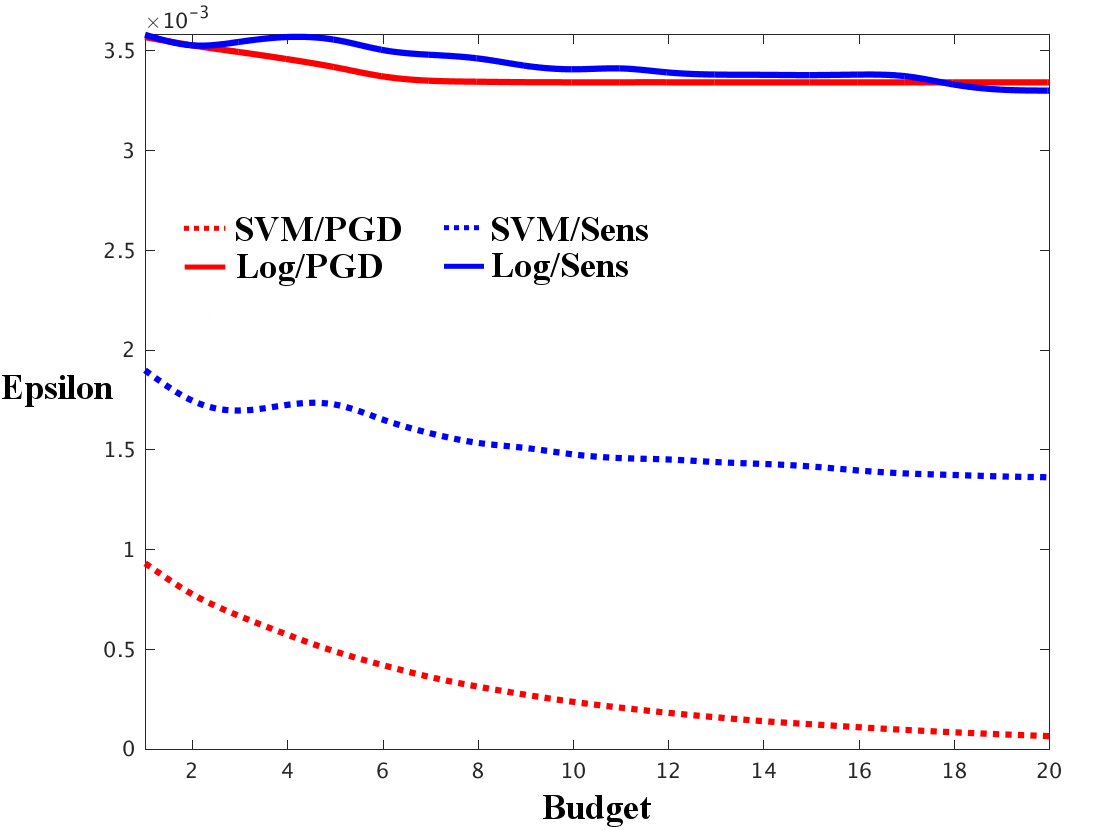}
		\caption{ARIC.~Average $\epsilon$ by budget.\label{fig:ar_eps_el}}
	\end{subfigure}
	\begin{subfigure}[]{.24\linewidth}
		\centering
		\includegraphics[scale=.09]{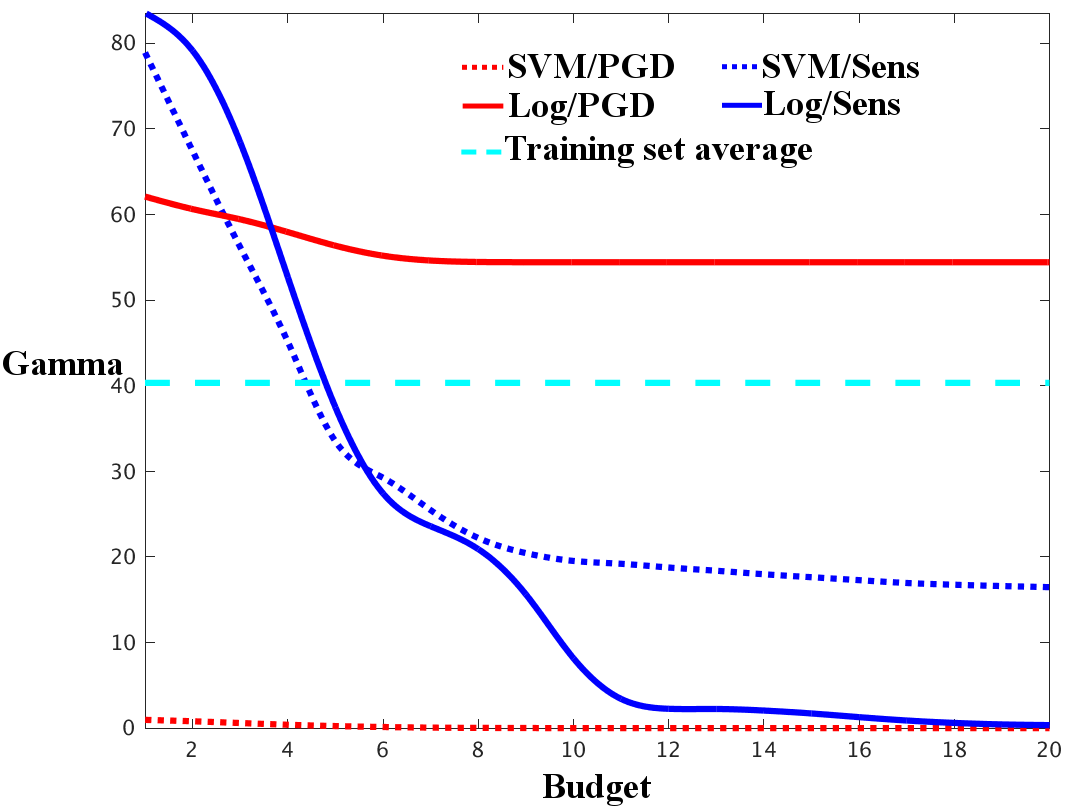}
		\caption{ARIC.~Average $\gamma$ by budget. \label{fig:hr_gam_el}}
	\end{subfigure}
	\caption{$(\epsilon,\gamma)$-support for Student Performance and ARIC using both the \textbf{Hardline} (\ref{fig:ben_eps_hl}-\ref{fig:ar_gam_hl}) and \textbf{Elastic} (\ref{fig:ben_eps_el}-\ref{fig:hr_gam_el}) bound-setting methods with $k=10$.\label{fig:eg_el}}
\end{figure*}

\subsection{Results: Cumulative and Individual Recommendations}

In this subsection we briefly relate the most common changes recommended to individuals in each dataset and then discuss the definitive recommendations made to two randomly selected instances.

Table \ref{tab:chg} shows the most common recommendations by raw count, the highest ranking of which pertain to features relevant to nearly all individuals (time with friends and eating food, for instance).

\begin{table}[h]
	\centering
	\begin{tabular}{|l|c|c|}
		\hline
		Rank & Student Perf. & ARIC \\
		\hline
		1 & Time w/ friends & Eat dark/grain bread \\
		\hline
		2 & Study time & Eat fruit\\
		\hline
		3 & Absences & \textcolor{red}{Cigs/day} \\
		\hline
		4 & \textcolor{red}{Weekday alco.~cons.} & Eat veggies\\
		\hline
	\end{tabular}
	\caption{Most commonly recommended feature changes by dataset using SVM with the PGD method at a budget of four. \label{tab:chg}}
\end{table}

Not all changes could be made to all individuals, however. For instance, not all individuals drink during the weekdays (Student Performance) and not all individuals smoke cigarettes(ARIC). Therefore, \textcolor{red}{red} shows that when recommendation commonality is normalized by the number of individuals who were engaging in weekday drinking and smoking, 97.97\% and 99.98\% of the time alterations to such behaviors were respectively recommended. Such a result shows that while such risky behaviors are not necessarily common among all individuals, those who do engage in them are frequently recommended to make alterations.

Figures \ref{fig:bench_kernel_rec} and \ref{fig:cvd_changes_hl} show the changes recommended to a randomly selected individual from Student Performance and ARIC, respectively, using SVM with the PGD method.


Contrasting Figure \ref{fig:bench_kernel_rec} with Figure \ref{fig:cvd_changes_hl} we can see that, in the case of the former, a single feature was optimized to the extent of feasibility before perturbations were made to another, whereas in the case of the latter, optimization of several features happened in tandem.

In examining the specific recommendations made to Student 135 in Figure \ref{fig:bench_kernel_rec}, we can see that first weekday drinking was curbed, followed by a reduction in school absences, weekend alcohol consumption, and time out with friends, as the budget was increased. Last, at the second highest budgetary level, time spent studying was increased. In the aggregate, it seems as though risk-related behavioral mitigations were determined to be optimal for this student.

Looking at the recommendations made to Patient 15 in Figure \ref{fig:cvd_changes_hl} we can see that, at low budgetary levels, an increase in dark or grain breads and a decrease in the number of cigarettes were recommended. Following these, as the budget was further incremented, consumption of more fruits and vegetables, in tandem, was recommended. At a budget of 13 it was also recommended that the patient decrease sodium intake and then subsequently, at a budget of 18, dietary fiber intake was increased. Finally, at a budget of 20, an increase in the consumption of nuts was recommended. Comprehensively, the recommendations deemed optimal for this patient were dietary-based, with the exception of a reduction in the number of cigarettes.

\subsection{Results: $(\epsilon,\gamma)$-support}

The results in Figure \ref{fig:eg_el} show that our inverse classifications are well supported in terms of probability space ($\epsilon$) and underlying training data ($\gamma$) for both Student Performance and ARIC, up to certain budgetary levels (sans SVM/PGD in \ref{fig:hr_gam_el}). This suggests that, in future work, a constraint on the underlying $\gamma$-support may be desirable. The results were obtained by taking the average over the $\epsilon,\gamma$ values of all optimized test instances for each budgetary level explored in past experiments.

\section{Conclusions}
%

In this work we propose and validate a new framework and method for inverse classification. The framework ensures that recommendations are realistic by accounting for what can actually be changed, the cost/effort required to make changes, the cumulative effort (budget) an individual is willing to put forth, and the effects that making changes have on features that are not directly actionable. Additionally, we impose bounds on the changeable features that further ensure recommendations are realistic, as well as two bound-setting methods that govern algorithmic recommendation-generating behavior. Furthermore, our methods are very modular, allowing for the use of any differentiable classification function (logistic regression, neural networks, etc.), as well as virtually any estimator of the indirectly changeable features. We demonstrated the efficacy of these methods on two freely available datasets as compared to a baseline method. Future work will focus on augmenting the framework with additional utility, as well as on conducting an in-depth analysis exploring situations in which PGD outperforms sensitivity analysis-based methods.

\bibliographystyle{IEEEtran}
\bibliography{lash_kdd_2017}

\section*{Appendix}

\subsection*{Proof of Proposition~\ref{prop:proj}}

Consider the index $i\in\mathcal{A}_-$. Due to the relationship $l_i'\leq z_i\leq u_i'\leq \min(0,w_i)$, any feasible value of $z_i$ can be at most $u_i'$ while deviating $z_i$ from $u_i'$ increases the objective value of \eqref{proj} and generates cost at a rate of $c_i^-$. Hence, the optimal value for $z_i$ must be $u_i'$ for each index $i\in\mathcal{A}_-$. Similarly, the optimal value for $z_i$ must be $l_i'$ for this index $i\in\mathcal{A}_+$.

With the optimal value of $z_i$ for $i\in\mathcal{A}_+\cup \mathcal{A}_-$ determined, the optimization problem \eqref{proj} is reduced to
\begin{eqnarray}
	\label{proj_reduce}
	\min_{\tilde\bz\in\Delta_{\tilde D}}&&\frac{1}{2}\|\tilde\bz-\tilde \bw\|^2
\end{eqnarray}
where $\tilde D= D\backslash(\mathcal{A}_+\cup \mathcal{A}_-)$, $\tilde \bw=\bw_{\tilde D}$, i.e., the sub-vector of $\bw$ containing the features in $\tilde D$, and
\small
\begin{align*}
	\label{FeasibleSetReduce}
	\Delta_{\tilde D}&\equiv&\left\{\tilde\bz\in\mathbb{R}^{|\tilde D|}\Bigg|
	\begin{array}{rl}
		&\sum_{i\in \tilde D}c_i^+(\tilde z_i)_++c_i^-(\tilde z_i)_-\\
		\leq& B-\sum_{i\in\mathcal{A}_-}u_i'c_i^--\sum_{i\in\mathcal{A}_+}l_i'c_i^+,\\
		&l_i'\leq \tilde z_i\leq u_i'\text{ for }i\in \tilde D.
	\end{array}
	\right\}.
\end{align*}
\normalsize
For any $\lambda\geq0$, let $z_i=\max\{\min\{h_i(w_i,\lambda),u_i'\},l_i'\}$ for $i\in\tilde D$. Using the definition of $h_i$ in \eqref{h}, we can show that the elements in the set
$$
z_i-w_i+\lambda c_i^+\partial(z_i)_++\lambda c_i^-\partial(z_i)_-
$$
are all positive only if $z_i=l'_i$ and the elements in the set
$$
z_i-w_i+\lambda c_i^+\partial(z_i)_++\lambda c_i^-\partial(z_i)_-
$$
are all negative only if $z_i=u'_i$ for any $i\in \tilde D$, where $\partial(z)_+$ and $\partial(z)_-$ represent the subdifferentials of the functions $(z)_+$ and $(z)_-$\footnote{Note that the subdifferential of a non-smooth function at some point can be a set.}. This indicates that $(z_i)_{i\in\tilde D}$ is the optimal solution of the Lagrangian relaxation problem
\begin{align*}
	\min_{l'_i\leq\tilde z_i\leq u'_i,i\in\tilde D}&&\frac{1}{2}\|\tilde\bz-\tilde \bw\|^2
	+\lambda\left(\sum_{i\in \tilde D}c_i^+(\tilde z_i)_++c_i^-(\tilde z_i)_-\right)
\end{align*}
with $\lambda$ being the Lagrangian multiplier. Step 4 and Step 8 in Algorithm~\eqref{algo:Proj} ensure $(z_i)_{i\in\tilde D}$ is a feasible solution of \eqref{proj_reduce} and satisfies the complementary slackness conditions with $\lambda$. This implies that $(z_i)_{i\in\tilde D}$ is the optimal solution of \eqref{proj_reduce} so that $(z_i)_{i\in D}$ is the optimal solution of \eqref{proj}.\qed

%
	
\subsection*{Proof of Proposition \ref{prop:0rep}}
Assume that the training set is drawn i.i.d.~from population distribution $\mathcal{P}$, having distribution $S$, where each dimension is in the range $[0,1]$, and that the size of the training set $n$ is large, then by the central limit theorem $\mu(S) {\xrightarrow {d}} \mu(\mathcal{P})\implies \text{disc}_{\mathcal{H}}(S,\mathcal{P}) {\xrightarrow {d}} \lVert \pmb{0}\rVert \leq \delta = 0$, as desired. \qed

\end{document}